# Real-Time Automatic Polyp Detection in Colonoscopy using Feature Enhancement Module and Spatiotemporal Similarity Correlation Unit

Jianwei Xu[a,†], Ran Zhao[b,†], Yizhou Yu[c], Qingwei Zhang[b], Xianzhang Bian[a], Jun Wang[a],

Zhizheng Ge[b,*], and Dahong Qian[a,*]

a.  Shanghai Jiao Tong University-Deepwise Healthcare Joint Research Lab, Institute of Medical Robotics, Shanghai Jiao Tong University, Shanghai, China

b.  Division of Gastroenterology and Hepatology, Key Laboratory of Gastroenterology and Hepatology, Ministry of Health, Renji Hospital, School of Medicine, Shanghai Jiao Tong University, Shanghai Institute of Digestive Disease, Shanghai, China

c.  Deepwise Artificial Intelligence Laboratory, Beijing, China

† These two authors contribute equally to the work.

*Corresponding author

Email address:

Zhizheng Ge, e-mail: zhizhengge@aliyun.com

Dahong Qian, e-mail: dahong.qian@sjtu.edu.cn

**Abstract:**

Automatic detection of polyps is challenging because different polyps vary greatly, while the changes between polyps and their analogues are small. The state-of-the-art methods are based on convolutional neural networks (CNNs). However, they may fail due to lack of training data, resulting in high rates of missed detection and false positives (FPs). In order to solve these problems, our method combines the two-dimensional (2-D) CNN-based real-time object detector network with spatiotemporal information. Firstly, we use a 2-D detector network to detect static images and frames, and based on the detector network, we propose two feature enhancement modules—the FP Relearning Module (FPRM) to make the detector network learning more about the features of FPs for higher precision, and the Image Style Transfer Module (ISTM) to enhance the features of polyps for sensitivity improvement. In video detection, we integrate spatiotemporal information, which uses Structural Similarity (SSIM) to measure the similarity between video frames. Finally, we propose the Inter-frame Similarity Correlation Unit (ISCU) to combine the results obtained by the detector network and frame similarity to make the final decision. We verify our method on both private databases and publicly available databases. Experimental results show that these modules and units provide a performance improvement compared with the baseline method. Comparison with the state-of-the-art methods shows that the proposed method outperforms the existing ones which can meet real-time constraints. It's demonstrated that our method provides a performance improvement in sensitivity, precision and specificity, and has great potential to be applied in clinical colonoscopy.

1. **Introduction**

Colorectal cancer (CRC) is the second leading cause of cancer death in the USA, with an estimated 147,950 new cases and 53,200 deaths in 2020 [1]. Some types (e.g., adenomatous polyp) of polyps are likely to develop into CRC, so the timely and accurate detection of polyps from optical colonoscopy videos is of great significance for the prevention and treatment of CRC. Traditionally, doctors use manual screening to look for polyps in colonoscopy videos, which can be time-consuming and inaccurate. It is reported that the missed detection rate of manual screening may be as high as 25% [2]. This high missed detection rate may lead to the late diagnosis of CRC and lower survival rates [3].

Thus, accurate real-time automatic detection of polyps in optical colonoscopy videos can be valuable in clinical practice, as it can free up time for doctors and save lives. However, this kind of detection is not easy to achieve, as polyps appear in varied shapes, sizes, colors, textures, and colorectal environments. Additionally, in the actual clinical examination, there may be jittering and blurring in the colonoscopy video due to manual operation.

## 1.1. Related Work

### 1.1.1. Polyp Detection

In early studies, hand-crafted features were extracted from polyp images and used to detect polyps. These features included color wavelet [4], texture [5], shape [6], local binary patterns [7,8] or a combination of them [9–11]. These hand-crafted feature methods used prior knowledge to delineate the features of polyps, then followed by a separate training process for classification or recognition.

In recent years, methods based on CNNs have been used in lesion detection and anatomical structure detection and have performed well [12–24]. This is due to the hidden feature representations provided by Convolutional Neutral Networks (CNNs) that show significant improvement over hand-crafted features. CNN-based methods use CNN to extract polyp features automatically, then followed by a classifier or detector to recognize polyps. These CNN-based detector networks can be mainly categorized as one- or multi -stage detectors. While the accuracy of the one-stage detector is not as good as that of the multi-stage detector, it has better performance in speed.

Multi-stage detectors use a region proposal network to find regions of interest for objects and then a classifier to refine the search to get the final predictions. Shin et al. [12] proposed to use the multi-stage detector Faster R-CNN architecture with a region proposal network and an inception-ResNet backbone to detect polyps. Mo et al. [17] and Sornapudi et al. [25] also proved that fine-tuning the pre-trained Faster R-CNN and Mask R-CNN can work considerably well in polyp detection. Moreover, Jia et al. [18] combined Faster R-CNN and Feature Pyramid Net for feature sharing, which is proven to be highly capable of guiding the learning process and improve recognition accuracy. However, these methods reported prediction speed was not suitable for real-time examination.

One-stage detectors perform a single pass on the data and incorporate anchor boxes to tackle multiple object detection such as YOLO-v3 [26] and SSD [27], which makes real-time object detection possible. The domain of Gastroenterology has also started to benefit from the success of single-stage object detectors. Urban et al. [19] and Zhang et al. [14] used YOLO to identify polyp regions in colonoscopy images in real-time. Zhang et al. [20] and Qadir et al. [15] proposed to use SSD to detect polyps from colonoscopy images in real-time.

In addition, the new anchor-free detectors are also receiving more and more research attention, such as CornerNet [28] and CenterNet [29]. Single and multi-stage detectors rely on a large number of preset anchors. Anchor free detectors detect objects through finding their cornet or center points by using classical backbones to generate a heatmap from the feature map showing potential spots of the object corners or center. Wang et al. [21] designed an anchor-free polyp detector which maintained real-time applicability.

**1.1.2. False positive reduction and true positive improvement**

CNN-based methods are limited by the lack of available labeled training data, especially in medical image processing, which may lead to low sensitivity and precision. In order to reduce FPs, based on Faster R-CNN, Shin et al. [12] proposed FP learning. They first trained a network with polyp images and generated FP samples with additional normal videos. They then retrained the network by adding back the generated FP samples.

Another approach to improving the performance is the consideration of temporal information. Yu et al. [13] proposed a three-dimensional (3-D) CNN network, whose input is a sequence of frames, to learn more contextual information [30,31] and introduced online learning to further reduce FPs. Shin et al. [12] used offline learning to learn video specific FPs. They first collect reliable polyp regions in the test video, then retrains the detector with these regions, and finally tests the video again with the newly trained detector. Angermann et al. [11] proposed the spatio-temporal coherence module. In this module, when calculating the final output for a given frame, the system also considered the outputs of the two previous frames. Zhang et al. [14] used a tracker after the 2-D detector network to increase sensitivity, but came at the cost of introducing more FPs. Qadir et al. [15] proposed the FP reduction stage, which integrated 7 previous and 7 future frames and detected irregularities and outliers using the coordinates of RoIs provided by a CNN-based detector.

**1.2. Our Contributions**

Our main contributions are summarized as follows:

1) We propose to combine the 2-D CNN detector network and spatiotemporal similarity to detect polyps.
2) Based on the image detector network, we propose two feature enhancement modules, 1) the FP relearning module (FPRM) and 2) the image style transfer module (ISTM), to further improve the precision and sensitivity of the detector network.
3) Then we propose the inter-frame similarity correlation unit (ISCU), which integrates spatiotemporal information, and combine it with the image detector network to improve performance in video detection.

## 2. Methods

As shown in Fig.1, our polyp detection system consists of two main parts: the CNN-based detector network and the inter-frame similarity correlation unit. To the CNN-based detector network, we add an FP Relearning Module and an Image Style Transfer Module during training to reduce false positives and improve sensitivity. This CNN-based detector network detects polyps frame by frame from a colonoscopy video. The outputs of this network are then fed into the second part of our system, the Inter-frame Similarity Correlation Unit. In this unit, several similar frames are correlated, and the results produced from the detector network are classified as true positives (TPs) or FPs and the rate of missed detection could be further reduced through the noise eliminating strategy and the missed detection correction strategy.

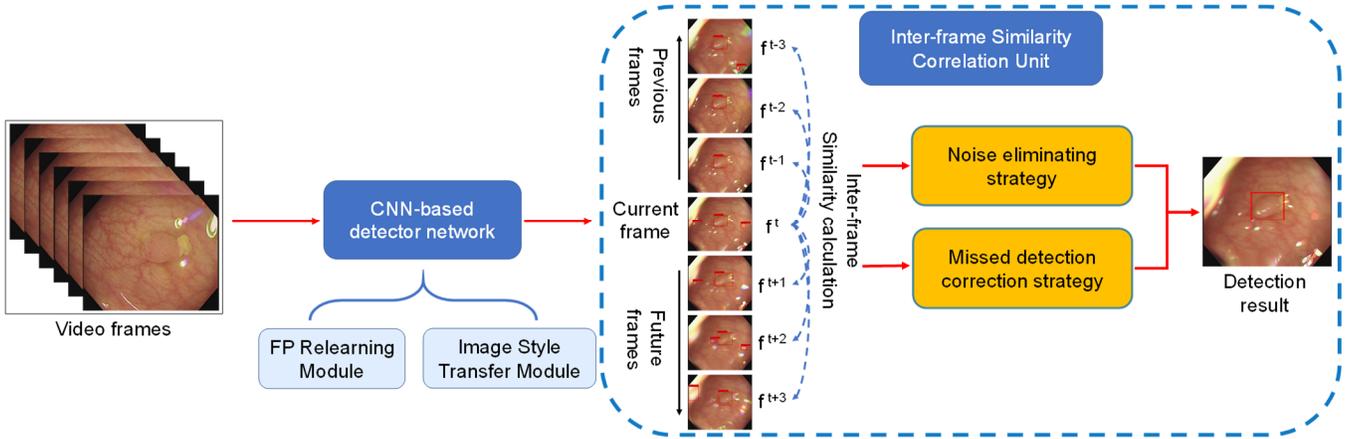

Fig. 1. Overview of our proposed method.

### 2.1. The CNN-based Detector Network

We first used a CNN-based object detection network in our system, which can directly regress to the bounding boxes of polyps. In our research, we used a one-stage detector, YOLOv3 [26], as the baseline network against which we could compare the performance of our method.

YOLOv3 divides the input image into $S*S$ grid cells. Each grid cell in the feature map predicts $B$ bounding boxes, and each bounding box includes center coordinates $(x, y)$, the height and width of the box $(h, w)$, the confidence score $c$, and the probability vector $p_k$ of classification. Finally, some of the redundant and overlapping bounding boxes are eliminated by non-maximum suppression (NMS). Each grid cell has $B$ anchor boxes, and the size of the boxes is preset using the K-means algorithm to cluster the representative shape, width, and height in the ground truth box of all samples in the training set. Then YOLOv3 calculates the Intersection over Union (IoU) for all anchor boxes, and the one that has the largest IoU with the ground truth box is a positive sample ($I_{ij}^{obj} = 1$) responsible for predicting the object. Other anchor boxes that have an IoU<0.5 with the ground truth are considered negative samples ($I_{ij}^{noobj} = 1$). The anchor boxes which do not have the best IoU and have an IoU>0.5 are ignored. The loss function is defined as follows:

$$\begin{aligned}\text{Loss} &= \sum_{i=0}^{s^2}\sum_{j=0}^{B} I_{ij}^{obj} * (2 - w*h) * [BCE(\sigma(x)) + BCE(\sigma(y))] \\ &+ \sum_{i=0}^{s^2}\sum_{j=0}^{B} I_{ij}^{obj} * (2 - w*h) * MSE(w,h) \\ &+ \sum_{i=0}^{s^2}\sum_{j=0}^{B} I_{ij}^{obj} * BCE(c) + \sum_{i=0}^{s^2}\sum_{j=0}^{B} I_{ij}^{noobj} * BCE(c) \\ &+ \sum_{i=0}^{s^2}\sum_{j=0}^{B} I_{ij}^{obj} \sum_{k=0}^{number\_class} BCE(p_k)\end{aligned} \quad (1)$$

### 2.1.1. The FP Relearning Module (FPRM)

To overcome the difficulty of reducing FPs, we propose the FP relearning module. The introduction of this module allows us to reduce FPs without the use of additional videos (Studies [13] and [19] used additional normal videos to generate FP samples, which could increase the cost of training and the difficulty of collecting data). First, we train an underfitting detector network using the original training set with only polyp samples, named "FP generator network". Next, we use the FP generator network to test the training set, and we find some FPs and their bounding boxes. We then add the FP bounding boxes as FP samples to the training data, and we call it "training set w/ FP samples". Lastly, we train a new detector network using the training data with both the original polyp samples and the added FP samples (i.e., training set w/ FP samples), and we call the new detector network "YOLOv3 w/ FPRM". The scheme is shown in Fig. 2.

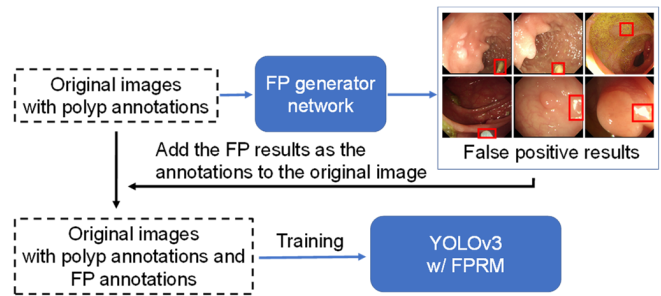

Fig. 2. The FP relearning module. It is used to generate specific FP samples to make the detector further learn the features of these specific false positives.

The key step of the FP relearning module is the generation of FP samples with which the image detector network is retrained. We want the FP samples to accurately represent actual FPs so that the image detector network is more capable of distinguishing polyps from polyp-like FPs.

In our study, we only keep the bounding boxes with confidence score>=0.3, and deem a bounding box to be an FP sample in case the bounding box is an FP result (IoU<0.5 with the ground truth). More specifically, FP samples with high confidence scores mean they may have features in common with polyps, while FP samples with low confidence scores mean they may be perturbations or noise.

Additionally, we need a balance between polyp samples and FP samples. The number of FP samples should not be too small, since if they were, the network may ignore the FP samples altogether and only focus on the original polyp samples. In order to generate enough FP samples, the FP generator network trained by the original training set with only polyp samples cannot be completely convergent (i.e., underfitting). It can achieve high sensitivity but low precision on the training data.

### 2.1.2. The Image Style Transfer Module (ISTM)

Results from Wickstrøm et al. [32] indicated that deep models are utilizing the shape and edge information of polyps to make better predictions. We thus wanted our detector network to learn more about polyps' edges and shapes while ignoring other features such as brightness, color, and texture.

Image style transfer [33] is used for rendering one image (the content image) into another style (the style image). And, we call this transferred image "the style-transferred image". Output image from this process retains the content of the original image but is simply in another style. We use this method of image style transfer to keep polyps' edge and shape information, while suppressing other features (color, brightness, texture, etc.). Adaptive Instance Normalization (AdaIN) [34] is used in this module. AdaIN consists of an encoder, an AdaIN layer and a decoder shown in Fig. 3. The AdaIN layer aligns the mean and variance of the content features with those of the style features to implement image style transfer. The AdaIN layer receives a content input $x$ and a style input $y$, and simply aligns the channel-wise mean and variance of $x$ to match those of $y$:

$$AdaIN(x, y) = \sigma(y)\left(\frac{x - \mu(x)}{\sigma(x)}\right) + \mu(y) \qquad (2)$$

where $\mu(\cdot), \sigma(\cdot)$ are the mean and standard deviation.

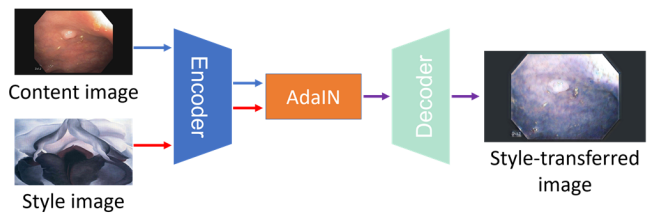

Fig. 3. An overview of AdaIN.

In our work, the encoder uses the first few layers (up to relu4_1) of VGG19 [35], and the decoder is the mirror of the encoder. Also, we replaced the pooling layers with the nearest upsampling layers. The entire AdaIN network is pretrained using MS-COCO [36] as content images and a dataset of paintings collected from WikiArt [37] as style images.

For each polyp image, we generate 11 style-transferred images, where the style images are randomly selected from the WikiArt dataset [37] and never observed during training. (We also tried different selections of the 11 style images, and more style images. But there is no obvious change to the result.) As shown in Fig. 4, this make sense, because the content of the original polyp images is retained, and we can still find the corresponding positions of the polyp in the style-transferred images. The 11 style-transferred images share the same annotation with the original polyp image, and all of them (style-transferred images and original polyp images) compose a new training dataset named "image-style-transferred training set". We use the image-style-transferred training set to train the detector network from the beginning, then fine-tune the network with the original polyp images dataset, and we call this new detector network "YOLOv3 w/ ISTM". The training scheme is shown in Fig. 5.

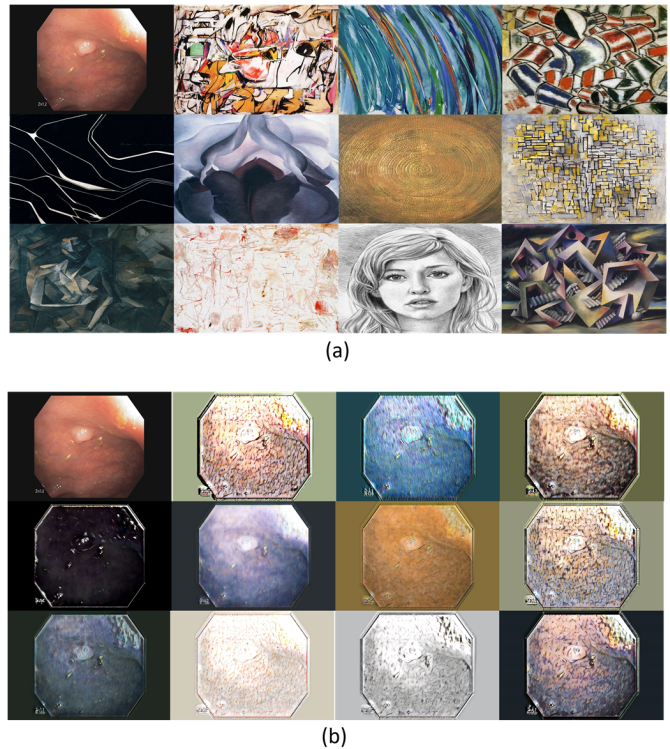

Fig. 4. (a) The content image (the top left image) and the style images. (b) The original image (top left image) and the style-transferred images.

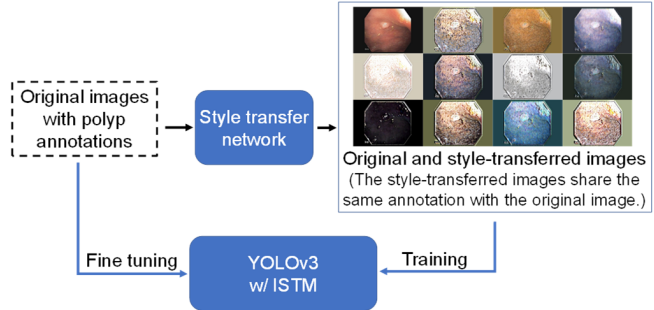

Fig. 5. The image style transfer module. It is used to enhance the edge and shape feature of polyps by generating style-transferred images.

### 2.1.3. The Combination of FPRM and ISTM

We first combine the FP samples and the style-transferred images generated above. More specifically, the 11 style-transferred images share the same annotation with training set w/ FP samples. It means the style-transferred images contain both polyp samples and FP samples generated above. The training set w/ FP samples and the style-transferred images with both polyp samples and FP samples consist a new training set we call it image-style-transferred training set w/ FP samples. We then train a new detector using the image-style-transferred training set w/ FP samples, and fine-tune the network with the training set w/ FP samples. We name the new detector network YOLOv3 w/ FPRM&ISTM. The scheme is shown in Fig. 6.

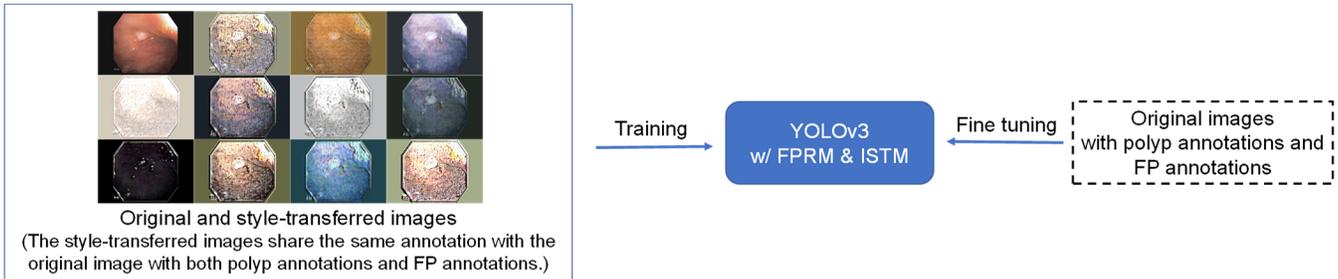

Fig. 6. The combination of FP relearning module and image style transfer module. We use the combination of FP samples and style-transferred images generated above to train the final detector network.

### 2.2. The Inter-frame Similarity Correlation Unit (ISCU)

Even with the addition of the FP relearning module, a system may still be prone to misjudging polyps in frames from a colonoscopy video. Our training dataset only includes clear polyp images, giving the detector the capability to distinguish FPs in clear images. However, detecting polyps in colonoscopy videos introduces new difficulties due to blurry frames, sudden changes of scene, light spots, overexposed regions, and the lens being too close to the intestinal wall. These noises are random and temporary, since most of the frames change smoothly. From this, we hypothesize that a polyp in one frame should appear in its neighboring frames at a similar location, while the noise cannot follow the smooth movement. Fig. 7 shows the movements of polyps and noise in seven frames. Frames 2 and 3 are clear, so the detector network correctly detects the polyp without any FPs. But frames 1, 4, 5, 6, and 7 are blurry, so the detector network generates several FPs.

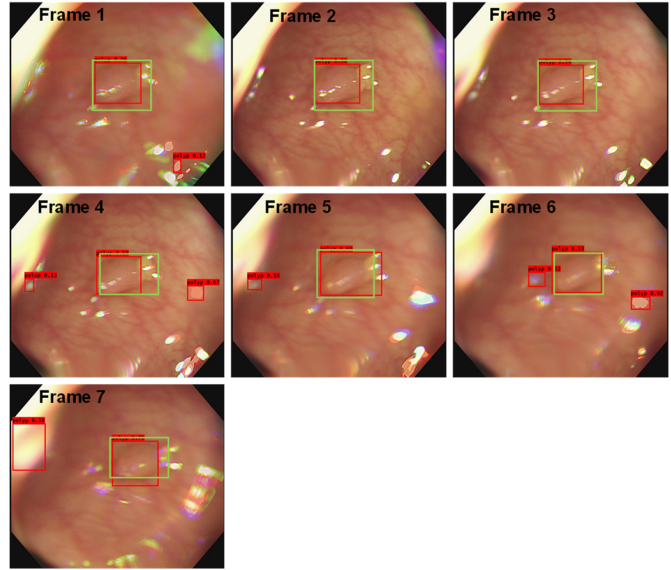

Fig. 7. A sequence of consecutive 7 frames. Green boxes represent the ground truth and red boxes represent the outputs of detector network. FPs appear in frame 1,4,5,6,7, but they are random and fast-moving.

We use spatiotemporal information to distinguish noise from the output bounding boxes and correct some of the missed detections due to the noise characteristics of randomness and transiency and polyp characteristic of continuity.

*1) Noise eliminating strategy*

To distinguish noise from polyps, we use the coordinates of the bounding boxes provided by the detector network and the similarity metrics of each neighbor frames as features. We use Structural Similarity (SSIM) [38] to measure the similarity between two frames.

SSIM compares two images ($x$ and $y$) using three statistics of image, including luminance $l$, contrast $con$ and structure $s$.

$$l(x,y) = \frac{2\mu_x\mu_y + b_1}{\mu_x^2 + \mu_y^2 + b_1} \quad (3)$$

$$con(x,y) = \frac{2\sigma_x\sigma_y + b_2}{\sigma_x^2 + \sigma_y^2 + b_2} \quad (4)$$

$$s(x,y) = \frac{\sigma_{xy} + b_3}{\sigma_x\sigma_y + b_3} \quad (5)$$

$$SSIM(x,y) = l(x,y) * con(x,y) * s(x,y) \quad (6)$$

where $\mu_x, \mu_y, \sigma_x^2, \sigma_y^2, \sigma_{xy}$ represent the mean, variance and covariance of $x$ and $y$, respectively. $b_1 = k_1 L^2$, $b_2 = k_2 L^2$ and $b_3 = b_2/2$ are three constants, avoid division by zero, where $L$ is the range of pixel value (commonly $L = 255$), and $k_1 = 0.01, k_2 = 0.03$ by default.

As shown in Fig. 1, some of the previous and future frames are considered in order to calculate the similarity metric with the current frame $f^t$. In our work, we only take 3 previous frames ($f^{t-3}, f^{t-2}, f^{t-1}$) and 3 future frames ($f^{t+1}, f^{t+2}, f^{t+3}$) into calculation. (The question regarding how many frames need to be considered will be discussed later in Section 4.4.) The detector network continuously outputs bounding boxes (confidence score>0.3) for each frame, and the number of bounding boxes in one frame is defined as $nb$. The coordinates $c$ of the bounding boxes in one frame form an array $a$, and all the arrays of the considered 7 frames form a set $S$, shown below.

$$S = \{a^{t-3}, a^{t-2}, a^{t-1}, a^t, a^{t+1}, a^{t+2}, a^{t+3}\} \quad (7)$$

$$a^{t+n} = [c_0, c_1, \ldots, c_{nb-1}] \quad (8)$$

$$n \in \{-3, -2, -1, 0, 1, 2, 3\} \quad (9)$$

$$c_i = [x_{min}^i, y_{min}^i, x_{max}^i, x_{max}^i] \quad (10)$$

$$i \in \{0, \ldots, nb - 1\} \quad (11)$$

We calculate the SSIM between the current frame $f^t$ and its six neighboring frames. For each of the $m$ frames that have a SSIM>0.85 with $f^t$, we compare its array of bounding boxes $a^{t+n}$ with the current frame's array $a^t$. A particular bounding box $c_i$ in the current frame's array $a^t$ is regarded as a TP if there is a bounding box overlapping with it (IoU>$IoU\_{th}$) in most of its similar frames (> $m/2$ frames). Otherwise, it will be treated as an FP, and will not be displayed in the output. The $IoU\_{th}$ is defined as:

$$IoU\_{th} = \frac{1}{2} * \left(\frac{x_{max}^i - x_{min}^i}{w_f} + \frac{y_{max}^i - y_{min}^i}{h_f}\right) \quad (12)$$

where $x_{min}^i, x_{max}^i, y_{min}^i, y_{max}^i$ represent the coordinates of the particular bounding box $c_i$ in the current frame's array $a^t$, and $w_f, h_f$ represent the width and height of the current frame. The reason of this definition is that the same movement between frames causes a large offset for small bounding boxes and a small offset for large bounding boxes. Therefore, for small bounding boxes, we allow them to have a smaller $IoU\_{th}$, while large bounding boxes are the opposite. So, we define an adaptive threshold here.

It could be that there is no similar frame (where SSIM>0.85) to the current frame among the 6 previous and future frames. In this case, we propose that a particular bounding box in the current frame is considered as a TP if there are at least 3 frames among the 6 that have an overlapping bounding box (IoU>*IoU_th*) with it. And we call this fixed-six-frames correlation (FC).

In Section 4.3, we also show the results of 1) our proposed method—using similarity metric to correlate similar frames while applying FC to those without similar frames, 2) applying FC directly to all frames without considering similarity metric. Through this, we demonstrate the effectiveness of involving similarity metric.

In this way, most of the FP bounding boxes caused by noise will be removed in the output, and only those bounding boxes that appear in adjacent or similar frames will be retained.

*2) Missed detection correction strategy*

In a collection of consecutive frames all containing the same polyp, a model may successfully detect the polyp in most of them but miss it in some of the frames. This could be due to fuzziness in the video, or the fact that CNNs are sensitive to small changes. Fig. 8 shows how the same polyp is detected successfully in frames 1, 2, 4, 5 and 6, but missed in frame 3.

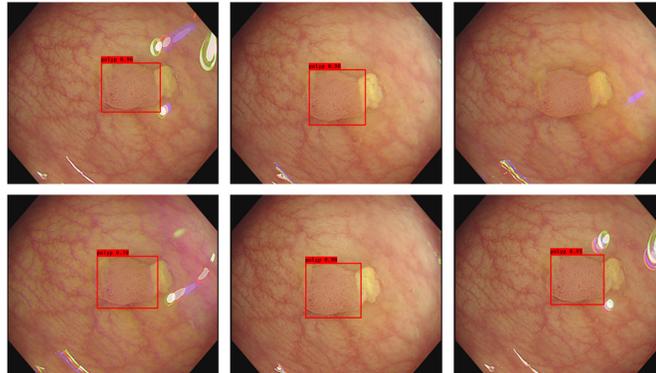

Fig. 8. A sequence of consecutive 6 frames. The polyp is missed by the detector network in frame 3.

In our method, we use an interpolation formula to add a bounding box around the missed polyp. More specifically, if there is no bounding box at a location in the current frame (the new bounding box doesn't have a IoU>0.5 with any bounding box in the current frame) but there is a bounding box at a similar location (IoU>0.5) in at least 3 of the 6 frames (these frames must contain both previous and future frames), we average the coordinates of these bounding boxes as a new coordinate $c$ and add it to the current frame:

$$c = \frac{1}{n}\sum_{i=1}^{n} c_i \qquad (13)$$

$$c_i = [x_{min}^i, y_{min}^i, x_{max}^i, x_{max}^i] \qquad (14)$$

where $n$ is the number of the frames which have a bounding box at a similar location and $c_i$ represents the coordinates of the bounding box.

These two strategies are applied separately on the original frames themselves. Noise elimination strategy determines whether the output bounding box from the previous detector network is FP or TP, and the FP box will be removed. For the new bounding boxes generated in correcting missed detections, all of them will be considered as TP, and will be output.

## 3. Experimental Setup

### 3.1. Experimental Datasets

In our study, we collected four publicly available datasets: CVC-ClinicDB (CVC-612) [39,40], CVC-ColonDB (CVC-300) [41], ETIS-LiribPolypDB [42] and CVC-ClinicVideoDB [11]. We also collected a static polyp image dataset and a colonoscopy video dataset from Renji Hospital affiliated to Shanghai Jiaotong University School of Medicine, respectively named RenjiImageDB and RenjiVideoDB. All datasets are summarized in Table 1 and Table 2.

Table 1
The Summary of Four Publicly Available Datasets and RenjiImageDB

|  | Dataset | Number of patient | Number of frames | Total number of polyps | Pixel resolution |
|---|---|---|---|---|---|
| PublicDB | CVC-ClinicDB | 23 | 612 | 646 | 388*284 |
|  | CVC-ColonDB | 13 | 300 | 300 | 500*574 |
|  | ETIS-LiribPolypDB | 34 | 196 | 208 | 1225*966 |
|  | CVC-ClinicVideoDB | 18 | 11954 | 10025 | 384*288 |
|  | Total | 88 | 13062 | 11179 | - |
| RenjiImageDB |  | 262 | 1482 | 1683 | 1200*966 |

Table 2
The Summary of RenjiVideoDB

| Index | Number of Frames | Total number of polyps | Number of positive frames | Number of negative frames | Polyp type |
|---|---|---|---|---|---|
| 1 | 38 | 38 | 38 | 0 | 0-Is |
| 2 | 162 | 162 | 162 | 0 | 0-Ip |
| 3 | 256 | 256 | 256 | 0 | 0-IIa |
| 4 | 120 | 120 | 120 | 0 | 0-Is |
| 5 | 149 | 149 | 149 | 0 | 0-IIa |
| 6 | 137 | 137 | 137 | 0 | 0-IIa |
| 7 | 361 | 361 | 361 | 0 | 0-Is |
| 8 | 123 | 123 | 123 | 0 | 0-Is |
| 9 | 1024 | 572 | 572 | 452 | 0-Is |
| 10 | 695 | 225 | 225 | 470 | 0-IIa |
| 11 | 2810 | 53 | 53 | 2757 | 0-IIa |
| 12 | 1251 | 439 | 439 | 812 | 0-Is |
| 13 | 542 | 635 | 492 | 50 | 0-Is |
| 14 | 1169 | 167 | 167 | 1002 | 0-Is |
| Total | 8837 | 3437 | 3294 | 5543 | - |

All of the publicly available datasets are annotated by clinical experts. The annotations provided for CVC-612, CVC-300 and ETIS-LiribPolypDB are exact boundaries that trace the polyp's edges, while the annotations for CVC-ClinicVideoDB are ellipses drawn around the polyps. In our study, we replace the original annotations with their minimum bounding rectangle which has four coordinates *x, y, h, w*. (*x, y*) are the centroid coordinates of the rectangle bounding box, and (*w, h*) are the width and height. For each video in CVC-ClinicVideoDB (because consecutive frames are highly correlated) we evenly extracted six frames containing polyps to form part of our training set.

RenjiImageDB consists of 1482 static polyp images from 262 patients with a pixel resolution of 1200*966. Each image contains at least one polyp, and there are 1683 polyps among 1482 images. RenjiVideoDB contains 14 videos from 14 patients, with a total of 8837 frames, of which 3294 frames contain polyps. Each video contains one individual polyp except video 11, while video 11 contains two individual polyps. All of them have a pixel resolution of 1280*1080. Details of the RenjiVideoDB are shown in Table 2. Both RenjiImageDB and RenjiVideoDB were collected during actual clinical examinations, so they serve well as test sets that can mimic actual clinical applications. Additionally, these datasets were annotated with rectangular boxes and verified by two experimental clinical experts. The polyps in RenjiVideoDB were also categorized based on Paris classification [43] by endoscopists. Paris classification buckets polyps into different types based on their morphology. This database contains the following three types: 1) 0-Ip—protruded pedunculated polyp in 1 video, 2) 0-Is—protruded sessile polyp in 8 videos, and 3) 0-IIa—superficial elevated polyp in 5 videos.

We used the four publicly available datasets, CVC-612, CVC-300, ETIS-LiribPolypDB and CVC-ClinicVideoDB (the latter of which 13 of the 18 videos were used, only six frames of each video), as our training set and named it PublicDB-Train. The remaining 5 videos in ClinicVideoDB were used as a validation set, named PublicDB-Val. All of the hyper-parameters mentioned in our work were optimized on PublicDB-Val to have the best F1-score. We used the RenjiImageDB and RenjiVideoDB datasets as the testing sets to verify our method.

### 3.2. Evaluation Metrics

We use common evaluation metrics of object detection to evaluate the performance of our method, which include Sensitivity (Sen), Precision (Pre), Specificity (Spe), F1-score (F1), F2-score (F2) and mAP. Our system outputs rectangular bounding boxes defined by four coordinates (*x, y, w, h*). One image can have one or more polyps and output bounding boxes, and all of the metrics are calculated on polyp level.

True Positive (TP): The output bounding box has an IoU>0.5 with one of the polyp ground truths. In the case the same polyp is detected multiple times, we only count one TP.

False Positive (FP): The output bounding box does not have an IoU>0.5 with any polyp ground truth.

False Negative (FN): There is no bounding box with an IoU>0.5 with a polyp's ground truth.

True Negative (TN): No output bounding box for a frame without a polyp.

Sensitivity, Precision and Specificity can be defined as follows, where TP+FN=the total number of polyps, and N is the number of frames without polyps:

$$Sen = \frac{TP}{TP + FN} \quad (15)$$

$$Pre = \frac{TP}{TP + FP} \quad (16)$$

$$Spe = \frac{TN}{N} \quad (17)$$

Moreover, to consider the balance between Precision and Sensitivity, we introduce the F1-score, F2-score and mAP.

$$F1 = \frac{2 * Sen * Pre}{Sen + Pre} \quad (18)$$

$$F2 = \frac{5 * Sen * Pre}{Sen + 4 * Pre} \quad (19)$$

Mean average precision (mAP): The area under the Precision-Sensitivity curve.

$$mAP = \int_0^1 Pre(Sen) \, dSen \quad (20)$$

For video detection, we also calculate the following metrics to account for clinical usability:

Polyp Detection Rate (PDR): Whether a method is able to detect the polyp at least once in a sequence.

Mean Number of False Positives per frame (MNFP).

Mean Processing Time (MPT)—the time needed for processing one frame.

$$MPT = \frac{\text{Total time to process one video}}{\text{Number of frames in this video}} \quad (21)$$

### 3.3. Training the Detector Networks

A small amount of training data is not enough to train an effective deep neural network [44]. In order to avoid the overfitting of the network and enhance the robustness of the network, we applied online augmentation strategies to each mini-batch of data, including random flipping, random changing aspect ratio and scaling, random image placing and color jittering.

In clinical colonoscopy videos, the same polyp may appear quite differently in different frames due to camera movement; size, shape, location and color may all vary frame to frame. We apply the augmentation strategies introduced to mitigate the effects of these variations. By changing the aspect ratio and scaling operations, the network could become more robust to variations in sizes and shapes. Adding the color jittering operation could make the detector network more robust to changes in color and brightness.

## 4. Experiments and Results

### 4.1. Analysis of FPRM and ISTM

In order to evaluate the performance of our method, we trained four detector networks, the baseline detector network YOLOv3, YOLOv3 with FP relearning module (FPRM), YOLOv3 with image style transfer module (ISTM), and YOLOv3 with both modules. All of them were trained on PublicDB-Train.

1) YOLOv3: we used the PublicDB-Train to train this baseline detector network. The learning rate was set as 0.001 initially and decreased by a factor of 10 every 200 epochs. The network was trained with 600 epochs, and we used the model weight with the lowest validation loss on the validation set.

2) YOLOv3-with-FPRM: we generated 188 FP samples following the scheme in Section 2.2. Then, we trained a new detector with the same setup as 1).

3) YOLOv3-with-ISTM: 1186 polyp images generated 13046 style-transferred images, and all 14232 images were used in training. First, we trained the network with the 14232 images with 200 epochs while the learning rate was set as 0.001. Then, we fine-tuned it using the original 1186 images with 400 epochs while the learning rate was set as 0.0001 initially and decreased by a factor of 10 every 200 epochs.

4) YOLOv3-with-FPRM-and-ISTM: We trained the detector using the 14232 images (with both polyp samples and FP samples) with the same setup as 3).

Table 3 shows the detection performance of the four detectors tested on RenjiImageDB. Table 4 shows how much each module improves the results. By comparing YOLOv3 with YOLOv3 w/ FPRM, and YOLOv3 w/ FPRM with YOLOv3 w/ FPRM&ISTM, we can find that the FP Relearning Module can reduce false positives and improve precision without reducing sensitivity. By comparing YOLOv3 with YOLOv3 w/ ISTM, and YOLOv3 w/ FPRM with YOLOv3 w/ FPRM&ISTM, we can find that the Image Style Transfer Module makes the detector more sensitive to polyps. YOLOv3-with-FPRM-and-ISTM has the highest sensitivity, F1-score, F2-score and mAP. The improvement of mAP means that the TP bounding boxes have higher confidence scores, while the FP bounding boxes have lower scores. By adjusting the confidence score threshold of the output, sensitivity from the high-mAP model will be higher than from the low-mAP model under the same precision. The Precision-Sensitivity curves are shown in Fig. 9.

On the other hand, through Table 4, we can find that the YOLOv3-with-ISTM and YOLOv3-with-FPRM-and-ISTM do not have an improvement in precision compared with YOLOv3. The reason may be that the style-transferred images may introduce some irrelevant features which belong to the style images. These features may cause problems for the network.

Table 3
Results on RenjiImageDB

Trained on PublicDB-Train and tested on RenjiImageDB

| Method | TP | FP | FN | Sen% | Pre% | F1% | F2% | mAP% | Sen% (Patients) |
|---|---|---|---|---|---|---|---|---|---|
| YOLOv3 | 1218 | 239 | 365 | 72.37 | 83.60 | 77.58 | 74.37 | 76.68 | 94.66 |
| w/ FPRM | 1212 | 166 | 371 | 72.01 | **87.95** | 79.19 | 74.72 | 78.60 | 95.03 |
| w/ ISTM | 1265 | 285 | 318 | 75.16 | 81.61 | 78.26 | 76.37 | 78.98 | 95.42 |
| w/ FPRM and ISTM | 1274 | 233 | 309 | **75.70** | 84.54 | **79.87** | **77.32** | **80.56** | **96.18** |

Table 4
Improvement with Each Module

| | YOLOv3 | | w/ FPRM | | w/ ISTM | | w/ FPRM and ISTM | |
|---|---|---|---|---|---|---|---|---|
| Improvement | Sen% | Pre% | Sen% | Pre% | Sen% | Pre% | Sen% | Pre% |
| YOLOv3 | - | - | -0.36 | +4.35 | +2.79 | -1.99 | +3.33 | +0.94 |
| w/ FPRM | - | - | - | - | - | - | +3.69 | -3.41 |
| w/ ISTM | - | - | - | - | - | - | +0.54 | +2.93 |
| w/ FPRM and ISTM | - | - | - | - | - | - | - | - |

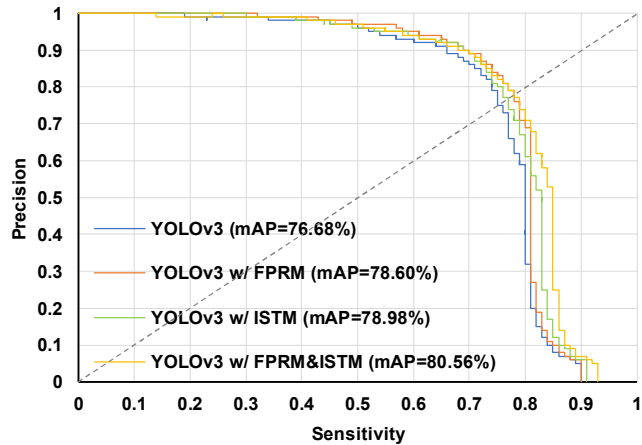

Fig. 9.  Precision-Sensitivity curves of the four detector networks.

### 4.2. Analysis of ISCU

In order to demonstrate the effectiveness of the inter-frame similarity correlation unit (ISCU), we showed the detection results of directly applying the detector network (trained on PublicDB-Train) to video detection and adding ISCU after the detector network. Table 5 presents the result obtained from the 14 videos from RenjiVideoDB. For the four detector networks listing in order in Table 5, with the integration of ISCU, the precision is improved by 22.99%, 17.18%, 21.04%, 16.12%, respectively, and the improvement of specificity is 26.99%, 15.59%, 23.89%, 12.45%, respectively. All of the polyps can be detected at least once in each video with ISTM (i.e., PDR=1).

Table 5
Results on RenjiVideoDB

| Method | Sen% | Improvement | Pre% | Improvement | F1% | Improvement | Spe% | Improvement | MNFP | PDR% |
|---|---|---|---|---|---|---|---|---|---|---|
| YOLOv3 | 80.39 | | 56.88 | | 66.62 | | 74.09 | | 0.24 | 93.33 |
| + FC | 79.95 | | 80.75 | | 80.35 | | 90.78 | | 0.07 | 93.33 |
| + ISCU | 80.83 | +0.44 | 82.68 | +25.80 | 81.74 | +15.12 | 92.04 | +21.56 | 0.07 | 93.33 |
| YOLOv3 w/ FPRM | 80.74 | | 67.78 | | 73.70 | | 80.08 | | 0.15 | 93.33 |
| + FC | 79.52 | | 83.96 | | 81.68 | | 92.39 | | 0.06 | 93.33 |
| + ISCU | 80.59 | -0.15 | 85.00 | +17.21 | 82.74 | +9.04 | 92.91 | +12.83 | 0.06 | 93.33 |
| YOLOv3 w/ ISTM | 85.10 | | 56.73 | | 68.08 | | 70.68 | | 0.25 | 100.0 |
| + FC | 84.46 | | 76.21 | | 80.13 | | 90.75 | | 0.10 | 100.0 |
| + ISCU | 85.63 | +0.52 | 78.17 | +21.44 | 81.73 | +13.65 | 92.12 | +21.43 | 0.09 | 100.0 |
| YOLOv3 w/ FPRM&ISTM | **87.55** | | 64.65 | | 74.38 | | 74.47 | | 0.19 | 100.0 |
| + FC | 86.76 | | 84.76 | | 85.75 | | 92.24 | | 0.06 | 100.0 |
| + ISCU | 87.49 | -0.06 | **85.99** | +21.33 | **86.73** | +12.35 | **92.93** | +18.46 | **0.06** | 100.0 |

From the results above, we demonstrate that the inter-frame similarity correlation unit plays an important role in video detection. Directly applying the 2-D CNN-based detector network to video detection causes a large number of FPs, because there are many blurred frames in the video, and the detector network have never seen these frames in the static clear image training dataset.

On the other hand, the results also show that the FP relearning module and the image style transfer module have an improvement on precision and sensitivity respectively. YOLOv3-with-FPRM-and-ISTM combined with ISCU has the best performance among the four detector networks.

### 4.3. Effect of Involving Similarity Metric

To know whether using SSIM as the similarity metric is useful in our method, we did an extra experiment to show the results of detector network only with fixed-six-frames correlation (FC) and detector network with inter-frame similarity correlation using SSIM.

Fixed-six-frames correlation means that we hypothesize that the current frame has correlation with all of the previous three frames and future three frames. And only the bounding boxes predicted in the current frame also appear within a similar position (IoU> $IoU_{\_th}$) in half of the six frames (>=3 frames) are further confirmed as true positives. It is also mentioned in case there are no similar frames with the current frame among the 6 previous and future frames. The missed detection correction strategy is the same as above.

The result in Table 5 shows that fixed-six-frames correlation results in a lower F1-score, because there may be a significant movement among the 7 consecutive frames in some cases. Also, the IoU of the two overlapping bounding boxes in two neighboring frames may be less than $IoU\_{th}$. If we fixedly correlate the 7 consecutive frames, some of the TP bounding boxes may be considered as FPs due to their fast movement. On other hand, one FP may repeat in half (3) of the 6 frames, but almost never repeats in most of the similar frames. In the inter-frame similarity correlation unit, we used SSIM to calculate the similarity metric first, and only correlate the similar frames. Therefore, we can eliminate most of the FPs and keep most of the TPs.

**4.4. Effect of the Number of Previous and Future Frames Considered**

To know how many frames we should consider in the inter-frame similarity correlation unit, we conducted an extra experiment: we considered $n/2$ previous frames and $n/2$ future frames, and where $n$ took on values from 2 to 16 in increments of 2. When $n$ is a small number, finding FPs may become difficult because it is possible that one individual FP repeats in a small number of frames. On the other hand, we may lose some TPs when $n$ is large, because TPs may be missed in certain frames of consecutive frames due to fuzziness in the video, or the fact that CNNs are sensitive to small changes. With more frames considered, more FPs were eliminated (Pre was improved) but more inconsecutive TPs were also be eliminated (Sen was degraded). Fig. 10 shows this trend.

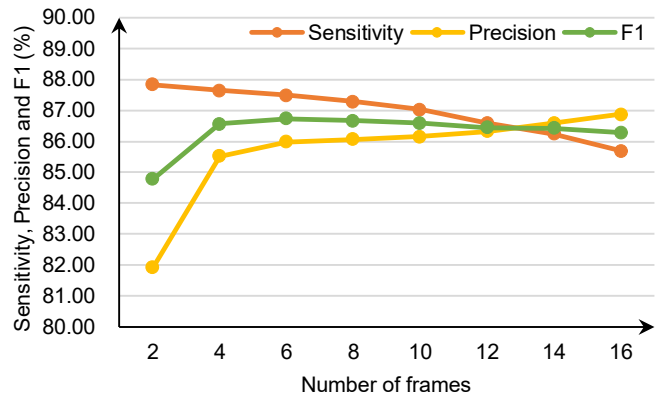

Fig. 10. Effect of consider n/2 previous frames and n/2 future frames, n changing from 2 to 16 in increments of 2. The results were obtained on RenjiVideoDB using YOLOv3-with-FPRM-and-ISTM.

**4.5. Performance in Different Types of Polyps**

Nonpolypoid (0-IIa, 0-IIb, 0-IIc) polyps are challenging to detect compared with polypoid polyps (0-Ip, 0-Is) and are often neglected until their potential to cause CRC is discovered [45]. So we also evaluated the performance in detecting different types of polyps. Polyps in RenjiVideoDB were categorized into three types including 0-Ip—protruded pedunculated polyp in 1 video, 0-Is—protruded sessile polyp in 8 videos, and 0-IIa—superficial elevated polyp in 5 videos, shown in Table 2.

Table 6 shows the results of our proposed method in detecting these three types of polyps. The protruded polyps (0-Ip and 0-Is) are easy to detect, because their edges and shapes features are quite obvious. The superficial elevated polyps are harder to detect because their edges and shapes features are difficult to distinguish. However, our proposed method has the capability to detect every individual polyp (i.e., PDR=1). Fig. 11 shows an example of each type of polyp.

Table 6
Results on Different Types of Polyps

A) 0-Ip protruded pedunculated polyp

| Method | Sen% | Pre% | F1% | Spe% | MNFP | PDR% |
|---|---|---|---|---|---|---|
| YOLOv3 | 100.0 | 91.53 | 95.58 | - | 0.09 | 100.0 |
| + ISCU | 100.0 | 91.53 | 95.58 | - | 0.09 | 100.0 |
| YOLOv3 w/ FPRM | 98.77 | 100.0 | 99.38 | - | 0.00 | 100.0 |
| + ISCU | 100.0 | 100.0 | 100.0 | - | 0.00 | 100.0 |
| YOLOv3 w/ ISTM | 99.38 | 93.06 | 96.12 | - | 0.07 | 100.0 |
| + ISCU | 100.0 | 92.05 | 95.86 | - | 0.09 | 100.0 |
| YOLOv3 w/ FPRM&ISTM | 100.0 | 98.78 | 99.39 | - | 0.01 | 100.0 |
| + ISCU | 100.0 | 100.0 | 100.0 | - | 0.00 | 100.0 |

B) 0-Is—protruded sessile polyp

| Method | Sen% | Pre% | F1% | Spe% | MNFP | PDR% |
|---|---|---|---|---|---|---|
| YOLOv3 | 90.75 | 67.35 | 77.32 | 72.63 | 0.23 | 100.0 |
| + ISCU | 90.92 | 84.74 | 87.72 | 87.78 | 0.08 | 100.0 |
| YOLOv3 w/ FPRM | 88.59 | 75.31 | 81.41 | 75.78 | 0.15 | 100.0 |
| + ISCU | 88.11 | 86.49 | 87.29 | 89.08 | 0.07 | 100.0 |
| YOLOv3 w/ ISTM | 97.35 | 69.48 | 81.09 | 67.14 | 0.22 | 100.0 |
| + ISCU | 97.56 | 85.44 | 91.10 | 89.64 | 0.09 | 100.0 |
| YOLOv3 w/ FPRM&ISTM | 96.95 | 76.09 | 85.26 | 73.27 | 0.16 | 100.0 |
| + ISCU | 97.11 | 90.65 | 93.77 | 91.93 | 0.05 | 100.0 |

C) 0-IIa—superficial elevated polyp

| Method | Sen% | Pre% | F1% | Spe% | MNFP | PDR% |
|---|---|---|---|---|---|---|
| YOLOv3 | 45.49 | 27.17 | 34.02 | 75.15 | 0.25 | 80.00 |
| + ISCU | 46.83 | 69.95 | 56.10 | 95.10 | 0.04 | 80.00 |
| YOLOv3 w/ FPRM | 53.66 | 42.07 | 47.16 | 83.17 | 0.15 | 80.00 |
| + ISCU | 54.27 | 74.66 | 62.85 | 95.66 | 0.04 | 80.00 |
| YOLOv3 w/ ISTM | 45.61 | 24.24 | 31.65 | 73.23 | 0.29 | 100.0 |
| + ISCU | 47.07 | 49.11 | 48.07 | 93.90 | 0.10 | 100.0 |
| YOLOv3 w/ FPRM&ISTM | 56.95 | 34.29 | 42.80 | 75.33 | 0.22 | 100.0 |
| + ISCU | 56.22 | 65.39 | 60.46 | 93.65 | 0.06 | 100.0 |

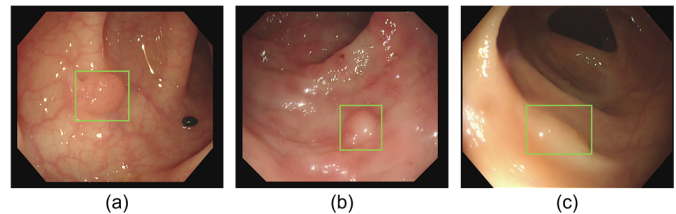

(a)     (b)     (c)

Fig. 11. Three types of polyps. (a) 0-Ip—protruded pedunculated polyp. (b) 0-Is—protruded sessile polyp. (c) 0-IIa—superficial elevated polyp.

### 4.6. Analysis of the Real-Time Performance

A polyp detection system also must process images at a minimum of 25 frames per second so as to be applicable during colonoscopy. In our experiment, the proposed method was implemented with Python under the deep learning framework Keras using a PC with a 3.6 GHz Intel(R) Core(TM) i7-9700 CPU and a NVIDIA GeForce RTX 2080TI GPU. To evaluate the processing time, we used the Mean Processing Time (MPT)—the time needed for processing one frame. MPT was 35 msec for the detector network and 2 msec for inter-frame similarity correlation unit, and the total MPT was 37 msec (i.e., 27 fps). In addition, there was delay caused by the inter-frame similarity correlation unit (120 msec), because the 3 future frames were considered to calculate the similarity. Therefore, our proposed method met real-time constraints (i.e., more than 25 fps).

### 4.7. Comparison with State-of-the-Art Methods

In Table 7, we compare the result of our proposed method with the results of others' previous work. All of the reported results are obtained from the corresponding references, and all of the methods used the same evaluation metrics presented in the 2015 MICCAI challenge [46] to perform fair evaluation.

Table 7(A) shows the still frame analysis results under the setting of 2015 MICCAI challenge [46]. All of the methods were trained on CVC-ClinicDB and tested on ETIS-LiribPolypDB. We only included the CNN-based methods: CUMED [46], OUS [46], Shin et al. [12], Sornapudi et al. [25], PLPNet [18], Haj-Manouchehri et al. [47]. CUMED [46] and Haj-Manouchehri et al. [47] used an end-to-end Fully Convolutional Network (FCN), taking an image as input and directly regresses the possibility that each pixel belongs to a polyp area. OUS [46] was based on a CNN-based classification network, and then used a sliding window to determine whether each sub-image belongs to the polyp area. Shin et al. [12] and Sornapudi et al. [25] respectively applied the Faster R-CNN and Mask R-CNN and fine-tuned the models with pre-trained weights on COCO datasets, and they also adopted several image augmentation strategies. PLPNet constructed a two-stage framework, which first detected potential polyp areas through a region proposal network (RPN), and then performed pixel-level segmentation through the FCN network.

Table 7(B) shows the video sequence analysis results. As we were not authorized to use the ASU-Mayo Clinic dataset, so we used CVC-ClinicVideoDB as our testing set and trained on CVC-ClinicDB (using the same setting as [11], [12] and [15]). Angermann et al. [11] extracted patches from the original images and calculated Local Binary Patterns (LBP) and Harr features of the patches, and used these features to train a classifier to determine whether the patches belong to the polyp region. Then they combined the positional relationship between the prediction boxes of the current frame and the previous two frames to give the final result of the current frame. Similarly, Qadir et al. [15] used the CNN-based object detection networks (i.e., Faster

R-CNN and SSD) to predict the regions where polyps may exist in each frame, and then combined the prediction results of the six frames before and after to give the final result of the current frame.

The results show that in both still frame analysis and video sequence analysis, our proposed method performs best in F1 score among all methods which can meet real-time constraints, and even performs better than most of the non-real-time methods. However, the F1 score of our method is not as high as Shin et al. [12]. Because they use Faster R-CNN which is a two-stage CNN-based detector network, and it can achieve a better accuracy but is more time-consuming (The MPT of Faster R-CNN is ten times that of our proposed method.). On the other hand, although we have better a F1-score than [18], [25] and [47], the sensitivity of our method is not as high as them. Because they use a segmentation-based method to find the centroid of polyps, and need exact boundaries of the polyp's edges during training, which allows the CNNs to learn more details. But this makes their methods have a higher false positive rate. At the same time, they need more time to process one frame, and need to spend more time on labeling the images before training. Our method only needs a roughly rectangular box of the region where the polyp is located, which greatly reduces the time required to label the images and speeds up the processing of one frame. This makes our method more valuable in clinical applications.

Table 7
Comparison with State-of-the-Art Methods

A) Still frame analysis. Trained on CVC-ClinicDB, tested on ETIS-LiribPolypDB

| Method | TP | FP | FN | Sen% | Pre% | F1% | F2% | MPT (msec) |
|---|---|---|---|---|---|---|---|---|
| CUMED [46] | 144 | 55 | 64 | 69.23 | 72.36 | 70.76 | 69.84 | 200 |
| OUS [46] | 131 | 57 | 77 | 62.98 | 69.68 | 66.16 | 64.22 | 5000 |
| Shin et al. [12] | 167 | 26 | 41 | 80.29 | 86.53 | 83.29 | 81.46 | 390 |
| Sornapudi et al. [25] | 167 | 62 | 41 | 80.29 | 72.93 | 76.43 | 78.70 | 317 |
| PLPNet [18] | 170 | 96 | 38 | 81.73 | 63.91 | 71.73 | 77.41 | 166 |
| Haj-Manouchehri et al. [47] | 183 | 94 | 25 | 87.98 | 66.06 | 75.46 | 82.51 | Not real-time |
| YOLOv3 w/ FPRM&ISTM (Ours) | 149 | 30 | 59 | 71.63 | 83.24 | 77.00 | 73.69 | 35 |

B) Video sequence analysis. Trained on CVC-ClinicDB, tested on CVC-ClinicVideoDB

| Method | TP | FP | TN | Sen% | Pre% | F1% | MPT (msec) |
|---|---|---|---|---|---|---|---|
| HarrN1 [11] | 4270 | 6650 | - | 42.59 | 39.10 | 40.77 | 21 |
| LBPN2+HarrN1 [11] | 5253 | 12026 | - | 52.40 | 30.40 | 38.48 | 185 |
| Faster R-CNN [12] | 8033 | 1648 | 1151 | 80.13 | 82.98 | 82.53 | 390 |
| Faster R-CNN+FP Reduction Unit [15] | 7904 | 829 | 1526 | 78.84 | 90.51 | 84.27 | 390 |
| SSD [15] | 5443 | 895 | 1629 | 54.29 | 85.88 | 66.53 | 33 |
| SSD+FP Reduction Unit [15] | 5329 | 399 | 1739 | 53.16 | 93.03 | 67.66 | 33 |
| YOLOv3 w/ FPRM&ISTM+ISCU (Ours) | 6653 | 862 | 1693 | 66.36 | 88.53 | 75.86 | 37 |

## 4.8. Hard cases visualization

We analyze some of the hard cases in RenjiVideoDB in Fig. 12, in which our method makes a failure prediction. Our predicted boxes are marked with blue rectangles, while the ground truth is marked with green rectangle. The first row shows the missed polyps (i.e., false negatives), and the second row shows the false positives.

We can see that most of the missed polyps are superficial elevated polyps (0-IIa), because their raised shapes are very inconspicuous, and their features are very similar to normal mucosae (i.e., Fig. 12(a)(b)(c)). However, although we missed these polyps in some frames, we successfully detected these polyps in most of the frames.

On the other hand, due to lighting and reflection issues, some overexposed or dark areas will appear during colonoscopy, which leads to FP results (i.e., Fig. 12(d)(e)). In addition, some of the FPs are due to polyp-like structures that are not actually polyps (i.e., Fig. 12(f)).

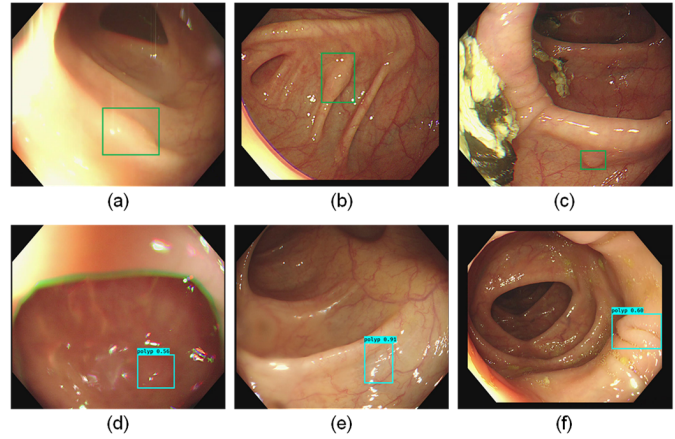

Fig. 12. Failure case of our proposed method on RenjiVideoDB. Ground truth is denoted by green rectangle and our prediction result is shown by blue rectangle. (a), (b) and (c) show the FN results caused by superficial elevated polyps. (d) and (e) show the FP results caused by dark and overexposed areas. (f) FP result caused by polyp-like raised structure.

5. Discussion

The original YOLOv3 [26] performs well in natural images, while polyp images have their own characteristics, such as less training data, similar image scenes, insufficient semantic information and so on. All these may cause poor detector network performance.

From the results of YOLOv3, we found that some FPs have similar features. For instance, light spots, bubbles, blood points, colon wall, and surgical instruments tended to be mistaken for polyps. We would like the detector network to learn the features of these specific FPs in order to avoid incorrectly detecting them as polyps. However, there are no FP samples in our dataset; instead, it consists only of images with polyps and their corresponding coordinates. As mentioned before, the detector network is trained with positive samples (the anchor box has the best IoU with the ground truth) and negative samples (the anchor boxes have an IoU<0.5 with the ground truth). It means the negative samples are fixed within the anchor boxes, and they cannot reflect the characteristics of the polyp-like FPs. The polyp-like FPs will also be mistaken for polyps in the output. Therefore, in the FP relearning module, we let the detector further learn the false positive results they produce. Unlike Shin et al. [12] and Qadir et al. [15], which generated FP samples from extra negative videos, we generated FP samples from the training set through an underfitting model, which means we needed less training data. Moreover, the use of FP learning in these previous studies [12] and [15] caused a degradation of sensitivity, but our proposed FP relearning module improves the precision without reducing the sensitivity.

It is demonstrated by Wickstrom et al. [32] that deep models are utilizing the shape and edge information of polyps to make their prediction. Based on this point of view, we innovatively propose the image style transfer module to enhance the edge and shape characteristics of polyps which could further improve the sensitivity of the detector network.

For video detection, our idea is similar to other real-time methods [14] and [15] in that we all integrate the one-stage detector and temporal information to achieve real-time and accurate polyp detection. In colonoscopy videos, most of the FPs are caused by light spots and blur due to camera shake (as shown in Fig.6). In our training dataset, only clear images are provided, so it was quite difficult for the detector network to adapt to blurred frames. However, these noises are random and transient, so they can be eliminated easily by integrating temporal information. We did not directly integrate the temporal information in the detector network, because the 3D-CNN network is quite time consuming (0.25s per frame) [13]. Zhang et al. [14] used a tracker after a real-time detector (i.e., YOLO) which only considered the previous frames. It is effective when there is a stable polyp per frame, however it may fail when there is severe jitter in the video or the polyp disappears in some frames. Similar to Qadir et al. [15], both of us integrated previous and future frames. They integrated 7 previous and 7 future frames fixedly based on the real-time detector (i.e., SSD). Differing from them, we first calculated the similarity between current frames and other frames, and only correlated the similar frames. Our results also showed that the inter-frame similarity correlation unit is greater than fixed-frame correlation.

However, our proposed method still has room for improvement in detecting flat polyps and eliminating polyp-like FPs. In this article, we only used the shallow information of the image to analyze the features of polyps and FPs. For instance, 1) in FP relearning module, we take those looks like polyps as FP samples, 2) in image style transfer module, we only enhance the edge and shape information of the polyps at the image level, 3) in inter-frame similarity correlation unit, we only use SSIM which is calculated based on pixel value to measure the similarity between frames. It is also important to the analyze the features in their latent space, such as the feature maps generated by CNN and optical flow features of videos. Because analyzing these feature maps can better explain directly why the neural network made such a prediction, and guide the framework to make a better prediction.

## 6. Conclusion

In this paper, we propose a novel framework for real-time polyp detection in colonoscopy, which gains accuracy from combining a CNN-based detector with two feature enhancement modules and a spatiotemporal correlation unit. The proposed method improves the polyp detection task by adding 1) the FPRM for relearning features of polyp-like false positives; 2) the ISTM which innovatively uses the image style transfer method to learn the edge and shape features of polyps. Moreover, enriching spatiotemporal information in video is also critical for polyp detection in colonoscopy. Hence, we proposed the ISCU to combine the structural similarity of neighbor frames and the results given by the previous detector network to give the final decision.

We evaluate the performance of our method on both private datasets and public datasets. Compared to the baseline model YOLOv3, our proposed method improves F1-score by 2.29% and 20.11% on RenjiImageDB and RenjiVideoDB, respectively. The proposed method also outperforms all of the real-time methods and even most of the non-real-time methods on public datasets. Additionally, for each individual polyp in video sequences, our method can successfully detect it in at least one frame (i.e., PDR=100%) and meet real-time constraints (MPT=37msec). Experimental results demonstrate the effectiveness of the proposed method and its potential practicability in clinical practice.

**Conflict of Interest**

The authors have no conflicts to disclose.

**Acknowledgment**

This work was supported in part by the National Natural Science Foundation of China under Grant 81974276, the Department of Science and Technology of Zhejiang Province-Key Research and Development Program under Grant 2017C03029. The authors would like to thank to the endoscopists in Renji Hospital affiliated to Shanghai Jiaotong University School of Medicine for their helpful contribution in collecting the colonoscopy datasets and providing the annotations.

Sanchez-Montes, S.R. Gurudu, G. Fernandez-Esparrach, X. Dray, J. Liang, A. Histace, Comparative Validation of Polyp Detection Methods in Video Colonoscopy: Results From the MICCAI 2015 Endoscopic Vision Challenge, IEEE Trans. Med. Imaging. 36 (2017) 1231–1249. https://doi.org/10.1109/TMI.2017.2664042.

[47] A. Haj-Manouchehri, H.M. Mohammadi, Polyp detection using CNNs in colonoscopy video, IET Comput. Vis. 14 (2020) 241–247. https://doi.org/10.1049/iet-cvi.2019.0300.